\documentclass[conference]{IEEEtran}
\IEEEoverridecommandlockouts
\usepackage{cite}
\usepackage{amsmath,amssymb,amsfonts}
\usepackage[noend]{algorithmic}

\usepackage{graphicx}
\usepackage{textcomp}
\usepackage{xcolor}

\usepackage{caption}
\usepackage{subcaption}
\usepackage{algorithm}

\usepackage{url}

\def\BibTeX{{\rm B\kern-.05em{\sc i\kern-.025em b}\kern-.08em
    T\kern-.1667em\lower.7ex\hbox{E}\kern-.125emX}}
\begin{document}

\title{Deceptive Level Generation for Angry Birds\\
}

\author{\IEEEauthorblockN{Chathura Gamage, Vimukthini Pinto, Jochen Renz}
\IEEEauthorblockA{\textit{School of Computing} \\
\textit{The Australian National University}\\
Canberra, Australia\\
\{chathura.gamage,u7069622,jochen.renz\}@anu.edu.au}

\and

\IEEEauthorblockN{Matthew Stephenson}
\IEEEauthorblockA{\textit{Department of Data Science and Knowledge Engineering} \\
\textit{Maastricht University}\\
Maastricht, the Netherlands \\
matthew.stephenson@maastrichtuniversity.nl}
}

\maketitle

\begin{abstract}
The Angry Birds AI competition has been held over many years to encourage the development of AI agents that can play Angry Birds game levels better than human players. Many different agents with various approaches have been employed over the competition's lifetime to solve this task. Even though the performance of these agents has increased significantly over the past few years, they still show major drawbacks in playing deceptive levels. This is because most of the current agents try to identify the best next shot rather than planning an effective sequence of shots. In order to encourage advancements in such agents, we present an automated methodology to generate deceptive game levels for Angry Birds. Even though there are many existing content generators for Angry Birds, they do not focus on generating deceptive levels. In this paper, we propose a procedure to generate deceptive levels for six deception categories that can fool the state-of-the-art Angry Birds playing AI agents. Our results show that generated deceptive levels exhibit similar characteristics of human-created deceptive levels. Additionally, we define metrics to measure the stability, solvability, and degree of deception of the generated levels.
\end{abstract}


\section{Introduction}
\label{introduction}

Procedural Level Generation (PLG) is a key area of investigation that focuses on the algorithmic generation of game levels in video game research \cite{mti1040027}, \cite{PCG_survey}. Game-playing agent development has benefited from PLG, since PLG can be used to generate a large number of training data in a short period of time \cite{DBLP:journals/corr/Togelius16}. Most of the current learning-based approaches embedded in AI agents require a huge amount of training data to be able to perform well \cite{zhu2016we}.

Video games are frequently used by AI researchers as testbeds for their research \cite{justesen2019we}. Angry Birds is one such example, for both PLG \cite{StephensonAIBIRDSLevel} and agent development \cite{renz2015aibirds}. Angry Birds is a physics-based puzzle game which provides interesting challenges for AI agents when solving the game levels. PLG is also non-trivial in Angry Birds due to the physics constraints in the game. Any PLG algorithm should adhere to the physical constraints of the game, and positions of the game objects should be determined with greater precision to ensure the expected outcomes can be obtained when playing. Similar to an agent in a physical environment, Angry Birds playing agents have a continuous action space. Hence, generating levels that can only be solved by intended actions is very difficult, especially when the levels are complex and require appropriate reasoning capabilities to solve.

A deception for an AI agent can be seen as a characteristic of a task which ``tricks'' the agent into making poor actions by exploiting its biases or limitations \cite{anderson2018deceptive}. A deceptive game level has a reward structure that can lead the agent away from the optimal strategy \cite{bontrager2019superstition}. Previous work on Angry Birds presented six categories of deceptions for the existing Angry Birds playing agents \cite{deceptive_angry_birds}. Using handcrafted levels for those deception categories, they showed the vulnerabilities of the state-of-the-art Angry Birds playing agents. The drawbacks of existing agents in handling deceptions show that there is room for further advancements of agent capabilities. This urges the necessity of creating a sufficient number of challenging deceptive levels to satisfy the data requirement of current learning approaches. 

In this paper, we present a methodology to generate deceptive levels in Angry Birds. We consider the six deception categories presented in \cite{deceptive_angry_birds} and define level templates that enable the automatic generation of varied levels for each of these deception categories. In addition to the level itself, a salient feature of this approach is that the solution for the level is also generated and can be utilized by the agents in the learning process. To evaluate our methodology, we define metrics to measure the stability, solvability, and deceptiveness of the generated levels. Moreover, to measure the characteristics of the generated levels compared to human-created levels, we evaluate AI agents' behaviour on both generated levels and handcrafted levels that require similar capabilities to solve.

\section{Background and Related Work} \label{related_work}

\subsection{Angry Birds}
Angry Birds is a 2D physics simulation game where players shoot birds from a slingshot to kill pigs. The game levels consist of dynamic objects such as birds, blocks, pigs, and TNT explosives that can move when forces are applied, and static objects such as platforms that are not affected by forces. Dynamic objects have health points that get reduced on collisions and they get destroyed and disappear when health points reach zero. There are three types of materials that blocks are made of: wood, stone, and ice. There are physical entities (i.e., structures) that are made up of various arrangements of these blocks. 
There are also five types of birds, some of which have special powers and strengths against certain materials:
\begin{itemize}
\item Red bird: No special power.
\item Blue bird: Splits into three birds when tapped, strong
against ice.
\item Yellow bird: Accelerates forward when tapped, strong against wood.
\item Black bird: Explodes when tapped, strong against stone.
\item White bird: Drops an explosive egg when
tapped.
\end{itemize}
The player cannot change the order of the birds given in a level. A player's action is a combination of 2 elements, 1) the release point of the bird from the slingshot 2) the tap time to activate the power of the bird, if available. A level is solved once all pigs are killed, using the provided birds. 
As the original Angry Birds game is not open-sourced, we used a research clone of Angry Birds developed in Unity \cite{ferreira2014search}. 


\subsection{PLG in Angry Birds}
Researchers have previously used several diverse approaches for procedurally generating game levels for Angry Birds \cite{ferreira2014search,jiang2017procedural,calle2019free,stephenson2017generating,abdullah2019angry}. The annual Angry Birds level generation competition \cite{StephensonAIBIRDSLevel} also promotes research into developing level generators for Angry Birds. This prior work mainly focuses on preserving the stability of the physical structures in the levels, adjusting the difficulty and enjoyability of the levels, and ensuring the levels are solvable. None of them focuses on generating deceptive game levels that require challenging reasoning and planning capabilities to solve. In this work, we focus on generating game levels that AI agents are not capable of solving through simple intuitive approaches. 

\subsection{Deceptive Games}
\label{Deceptive Games}
The concept of deceptive games was first presented in the study by Anderson et al. \cite{anderson2018deceptive}. They create a suite of deceptive arcade-style games for the General Video Game AI (GVGAI) framework \cite{perez2016general} and explore the effect of those games on game-playing agents. Building on that idea, an approach to generate deceptive levels for games in the GVGAI platform is discussed in \cite{zafar2018deceptive} and a methodology to evaluate agents on deceptive levels is presented in \cite{sabahevaluating}. 

\begin{figure}[t]
  \begin{subfigure}[b]{0.49\columnwidth}
    \includegraphics[width=\linewidth]{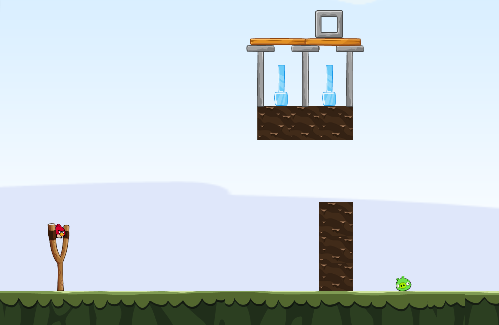}
    \caption{Rolling/falling objects}
    \label{rolling_with_sol}
  \end{subfigure}
  \hfill 
  \begin{subfigure}[b]{0.49\columnwidth}
    \includegraphics[width=\linewidth]{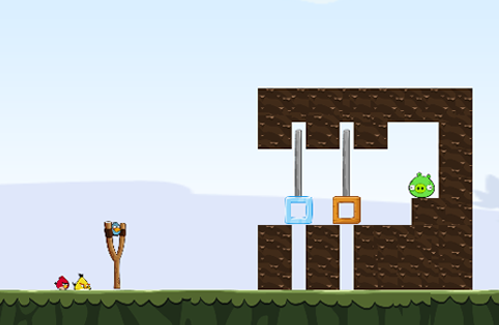}
    \caption{Clearing paths}
    \label{clearing_with_sol}
  \end{subfigure}
    \begin{subfigure}[b]{0.49\columnwidth}
    \includegraphics[width=\linewidth]{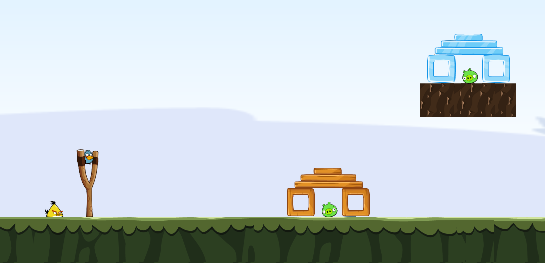}
    \caption{Entity strength analysis}
    \label{strength_with_sol}
  \end{subfigure}
  \hfill 
  \begin{subfigure}[b]{0.49\columnwidth}
    \includegraphics[width=\linewidth]{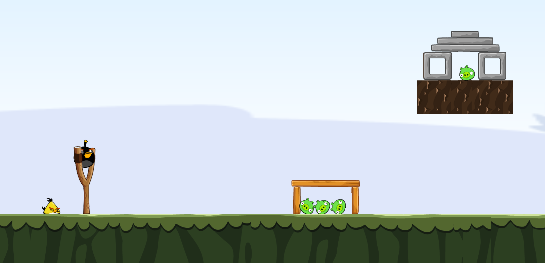}
    \caption{Non-greedy actions}
    \label{greedy_with_sol}
  \end{subfigure}
  \begin{subfigure}[b]{0.49\columnwidth}
    \includegraphics[width=\linewidth]{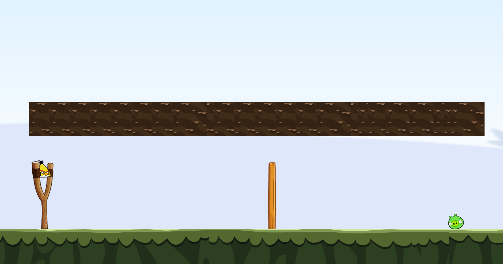}
    \caption{Non-fixed tap time}
    \label{greedy_with_sol}
  \end{subfigure}
  \begin{subfigure}[b]{0.49\columnwidth}
    \includegraphics[width=\linewidth]{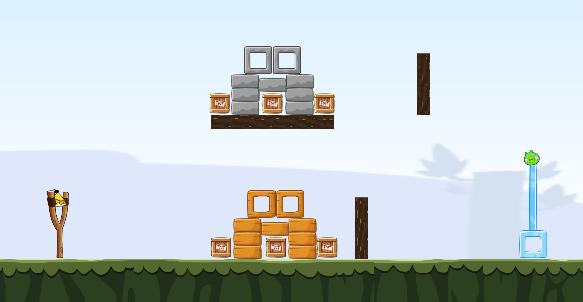}
    \caption{TNT}
    \label{greedy_with_sol}
  \end{subfigure}
\caption{Six example handcrafted levels for each deception category presented in \cite{deceptive_angry_birds}. The solutions for the levels are, (a) target the structure which collapses and falls on top of pig, (b) use first two birds to clear path for third, (c) use blue and yellow birds respectively on ice and wood structures, (d) make non-greedy shot that kills less pigs with black bird, (e) tap the yellow bird (accelerate) before hitting the block, and (f) target pig directly and ignore TNTs.}
\label{examples_for_4_deceptions_hand_crafted}
\end{figure}

In the context of complex physics simulation games, as already mentioned in Section \ref{introduction}, the work on deceptive Angry Birds levels by Stephenson and Renz \cite{deceptive_angry_birds} suggests six different categories of deceptions that can trick or exploit the current state-of-the-art Angry Birds playing agents. While these deceptions do not affect all Angry Birds agents equally, no agent was able to successfully handle all deception categories. We consider these same six deception categories in our generation process. The six deception categories are described below and examples for those categories are shown in Fig. \ref{examples_for_4_deceptions_hand_crafted}.

\subsubsection{Rolling/falling objects}
This deception uses the fact that objects of one entity can fall or be rolled on to another entity to create an impact. An agent needs to understand that an object can fall or be rolled and that object can be used to hit a target.

\subsubsection{Clearing paths}
This deception occurs when there are obstacles that need to be removed or destroyed before a target can be reached. To deal with this deception, an agent needs to understand that in order to reach a target, it should first clear the path to the target.

\subsubsection{Entity strength analysis}
This deception requires an agent to analyse the strength of the entities in a level. The strength of an entity depends on the factors such as its material, shape and size. An agent should be capable of determining the physical weaknesses/strengths of individual entities and interact accordingly to solve the level.

\subsubsection{Non-greedy actions}
In this deception, an action that appears to be less effective in the short term compared to another possible action, will gain higher advantages in the long term. The agent should look ahead and plan the actions using its knowledge of the environment rather than performing a greedy action that gives a higher short term reward.

\subsubsection{Non-fixed tap time}
The special powers of the birds are activated by tapping while the bird is in flight. In this deception, an agent needs to activate the special powers of the birds at non-fixed times, opposed to tapping at a pre-determined fixed time.

\subsubsection{TNT}
The TNT explosives in the game explode when hit, causing damage or pushing nearby objects. In this deception, an agent needs to use TNTs to kill pigs or TNTs are used to distract the agent from the objective of killing pigs.

\begin{figure*}[t]
\centerline{\includegraphics[width=1\textwidth]{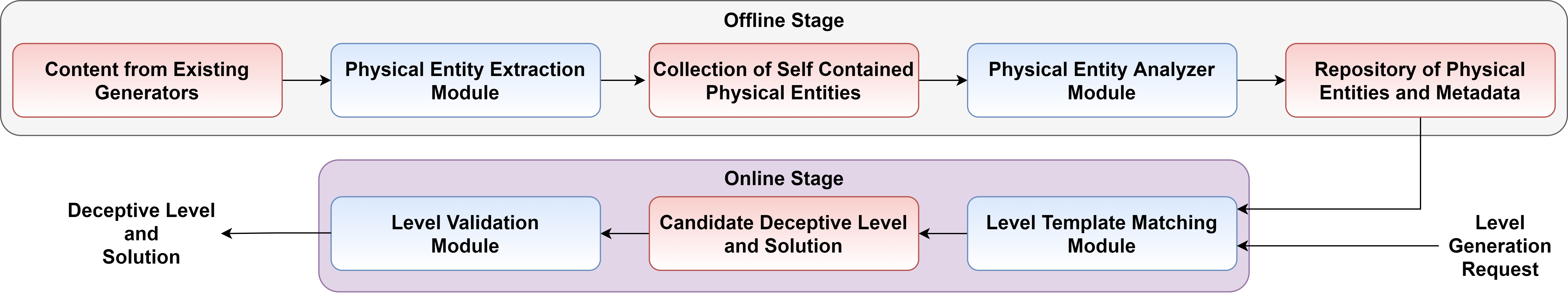}}
\caption{Deceptive level generation methodology.}
\label{generation_procedure}
\end{figure*}

\section{Terminology} \label{terminology}

In this paper, we use the term {\it environment} to refer to the space where an agent can sense and act, which is not a part of the agent. 
A stand-alone, self-contained set of physically stable objects in the environment is termed as a {\it physical entity}. A {\it deception} is considered as a characteristic of a task that exploits the cognitive biases of an agent and causes it to make sub-optimal decisions \cite{anderson2018deceptive}.
A {\it strategy} is a sequence of an agent's actions (i.e., bird release points and tap times) that involves interacting with the environment. A {\it solution strategy} is a specific strategy that solves a given level when executed. Finally, a {\it tactic} is a plan that an agent tries to solve a level (e.g. shooting birds targeting at pigs). For a tactic, there can be multiple possible strategies which may or may not solve the level. 

\section{Proposed Procedure} \label{propsed_procedure}

In this section, we present the proposed deceptive level generation procedure for Angry Birds. Fig. \ref{generation_procedure} shows the main components of the generation procedure. The four components shown in blue are the modules in the procedure and the four components in red are the inputs/outputs of those modules. The generation procedure does not involve creating physical entities; instead, it extracts physical entities created from existing content generators. This provides access to a wide range of content from various generation methods. Similarly to using existing content generators, customized handcrafted content can be used as well. The generation procedure consists of two stages: an offline stage and an online stage. The offline stage is shown inside the grey box (first row) and the online stage is shown inside the purple box (second row). The offline stage needs to be executed only once for an extracted set of physical entities. When a level generation request is received, the online stage is executed to generate a level using the offline stage's stored data. In subsequent subsections, we discuss the four components in our generation procedure: \textit{Physical Entity Extraction Module}, \textit{Physical Entity Analyzer Module}, \textit{Level Template Matching Module}, and \textit{Level Validation Module}.

\subsection{Physical Entity Extraction Module}

\begin{figure}[t]
\centerline{\includegraphics[width=0.47\textwidth]{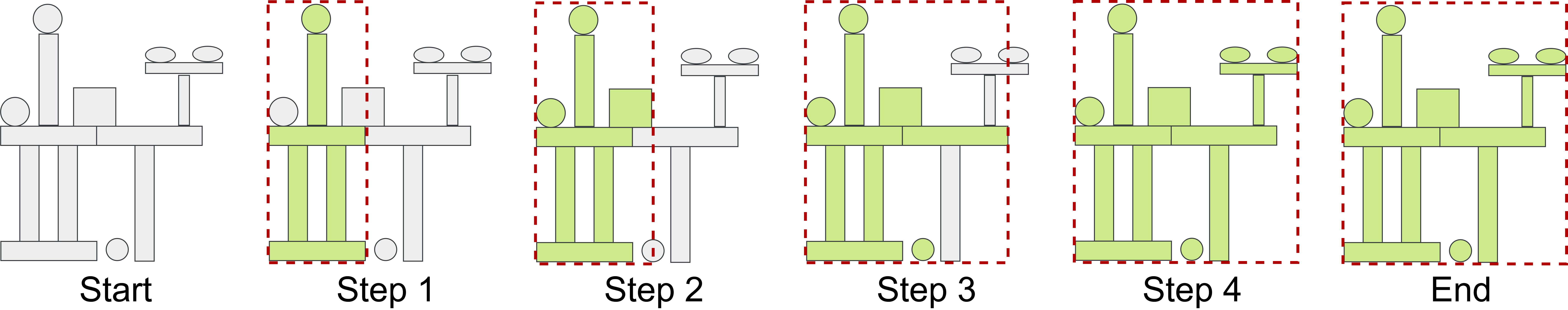}}
\caption{An illustration of the extraction process for a single entity. The leftmost figure (start) shows the entity that needs to be extracted. Each step corresponds to an iteration of the algorithm and the rightmost figure (end) shows the extracted entity. Selected objects in each step are shown in green, the bounding box of selected objects are shown in red dotted lines.}
\label{entity_extraction_process}
\end{figure}

This module takes already generated game instances from Angry Birds content generators and extracts physical entities from those instances. The game instances can be complete game levels or parts of game levels with physical entities. We use a collection of existing Angry Birds level generators \cite{StephensonAIBIRDSLevel} to generate a set of game instances. Entities are copied from these instances using a qualitative reasoning process that iteratively extracts entities. Algorithm \ref{alg:algorithm_entity_extraction} shows the physical entity extraction process that can be applied to a game instance with multiple physical entities. When formulating the algorithm, we define the term supporter using the support graph similar to how authors have defined it in \cite{article}. If the bottom horizontal edge of an object $O_i$ is in contact with the top horizontal edge of another object $O_j$ (i.e., $O_i$ is resting on top of $O_j$) then the support graph contains an edge pointing from object $O_i$ to $O_j$. In a given support graph, if there exists a path from object $O_i$ to object $O_j$, then object $O_j$ is considered as a supporter of the object $O_i$ ($O_j$ supports $O_i$). Objects placed on the ground do not have any supporter and hence has an empty support graph. Fig. \ref{entity_extraction_process} shows an example of how a physical entity is extracted iteratively using this algorithm.

\begin{algorithm}[t]
\footnotesize
\algsetup{linenosize=\footnotesize}
\caption{Extract physical entities}
\label{alg:algorithm_entity_extraction}
\textbf{Input}: Game instance with physical entities\\
\textbf{Output}: Set of extracted physical entities
\begin{algorithmic}[1] 
\STATE entitiesExtracted = \{\}
\STATE objectsRemaining = all objects in the instance
\WHILE{objectsRemaining is not empty}
\STATE topObject = topmost object of objectsRemaining
\STATE entityExtracting = topObject and supporters of topObject
\WHILE{True}
\STATE boundingBox = bounding box of entityExtracting
\STATE entityExtracting = objects inside boundingBox
\IF {boundingBox size not increased}
\STATE Break
\ENDIF
\ENDWHILE
\STATE add entityExtracting to entitiesExtracted
\STATE remove objects of entityExtracting from objectsRemaining
\ENDWHILE
\STATE \textbf{return} entitiesExtracted
\end{algorithmic}
\end{algorithm}

\subsection{Physical Entity Analyzer Module}\label{entity_analyzer_module}

After extracting physical entities, the \textit{Physical Entity Analyzer Module} is used to analyze those entities individually by interacting with them and observing the outcome. 
In Angry Birds, interactions with the entities can only be done by shooting birds from the slingshot and tapping to activate the bird's special power. To predict the result of physical interactions we can use either qualitative methods \cite{WalegaQualitative,Peng2014}, which are typically faster but less accurate, or simulation-based methods \cite{Battaglia18327}, which are typically more accurate but slower. Using qualitative methods in Angry Birds for physical predictions has proven to be less accurate and robust for large complex entities \cite{stephenson2017generating}. Therefore, we use a simulation-based method to perform interactions and record the outcomes. 

We developed a portfolio agent which can play Angry Birds with four different tactics adopted from previous Angry Birds playing AI agents \cite{stephenson20182017}. They are, 1) shooting birds (without special powers) targeting pigs in the entity, 2) shooting birds (without special powers) targeting at TNTs in the entity, 3) shooting birds (without special powers) targeting at reachable blocks in the entity, and 4) shooting birds with special powers (activated at different times) targeting at pigs.
The above tactics also have different variants (i.e., strategies) based on the order of the targets (if multiple targets exist), the shooting angle, and the activation time of the bird special power. 

Each entity is tested with a predetermined fixed set of strategies. When interacting with the entities, the objects in the entity can be subjected to move, collide, damage, or destroy effects, due to the force applied on the entity. For each interaction (i.e., a bird shot), data is recorded systematically in four steps.
\begin{itemize}
    \item Entity data before the interaction (e.g. the number of objects in the entity, the entity's bounding box size).
    \item Interaction data (e.g. the magnitude and the location of the force applied on the entity due to the interaction).
    \item Dynamic data when the objects in the entity are moving as a result of the interaction (e.g. objects that have gone out of the entity's original bounding box, location and the velocity of the objects gone out). 
    \item Entity data after the interaction when all the objects become stationary (e.g. the number of objects in the entity, the entity's bounding box size).
\end{itemize}
The data gathered through this analysis is attached to the entity as metadata and a repository of entities with metadata is maintained. This is the end of the offline stage of the generation process, the next steps are done online.

\subsection{Level Template Matching Module} \label{deceptive_task_template_matching}

The online stage of the physical deceptive level generation process starts with a generation request coming to the \textit{Level Template Matching Module}, specifying the desired deception category. This module generates a candidate level for that deception category using pre-defined level templates. 
For a deception, a level template contains constraints that entities should satisfy and rules for level generation. When generating a level, the level template considers possible interactions that an agent can perform and the outcomes of those interactions obtained from the entities' metadata. The entities that satisfy the constraints in the template are used to generate the level, following the generation rules.

When generating levels, \textit{Level Template Matching Module} also creates the solution for the level using the interaction data (bird shots) in the metadata of the entities used to generate the level. The solution has the interaction data that solves the level (i.e., the solution strategy of the level). When designing the level templates, we attempt to make the deceptive level solvable only by using the generated strategy. We assume if an agent uses the solution strategy to solve the level, the agent understood the deception. Restricting the solvability only to the generated solution prevents the agents from solving the levels using other tactics without realizing the deception. 

To facilitate the template design discussion, on top of the terminology discussed in Section \ref{terminology}, we define the following terms. The \textit{goal} of an Angry Birds player is, killing all the pigs in the level using the given number of birds. We refer to the term \textit{solving an entity} as killing all the pigs within an entity. An entity is deemed solvable if all the pigs within the entity can be killed using the given birds. An \textit{outperforming solution strategy (OSS)} refers to the solution strategy that uses the minimum number of birds to solve an entity compared to other solution strategies tested. If there are multiple solution strategies with the fewest number of birds needed to solve an entity, then there is no OSS for that entity. The level templates designed for the six deception categories are explained in the paragraphs below.

\subsubsection{Rolling/Falling Objects}
A level with this deception is created using two types of entities, called senders and receivers. A sender gives an object out either by rolling or falling when an agent interacts with the entity. A receiver uses the impact of the sender's rolling/falling object towards achieving the goal (i.e., killing the pigs). The locations of the two entities are determined such that the rolling/falling object's impact can be used to replace the impact of an action in the receivers' OSS. The placement of the two entities should also ensure that the above-considered action no longer allows the agent to use the original strategies for solving the level without understanding the deception. The reachability of the target objects for the birds shot from the slingshot is verified to ensure the solvability of the level. 
Algorithm \ref{alg:algorithm_rolling_falling} shows the pseudocode of this level template.

\begin{algorithm}[t]
\footnotesize
\algsetup{linenosize=\footnotesize}
\caption{Rolling/falling objects}
\label{alg:algorithm_rolling_falling}
\textbf{Input}: Sender entity, Receiver entity\\
\textbf{Output}: A level with rolling/falling objects deception

\begin{algorithmic}[1] 
\IF {sender has suitable objects that can be rolled/fallen}
\IF {receiver has an OSS}
\WHILE{generation unsuccessful}
\STATE get sender's rolling/falling object's trajectory
\STATE get OSS of receiver
\STATE generate level by matching sender and receiver
\STATE verify reachability of targets
\IF {generation successful}
\STATE \textbf{break}
\ENDIF
\IF {maximum generation attempts reached}
\STATE \textbf{return} none
\ENDIF
\ENDWHILE
\STATE allocate birds to level
\STATE generate solution strategy
\ENDIF
\ENDIF
\STATE \textbf{return} generated level
\end{algorithmic}
\end{algorithm}

\subsubsection{Clearing Paths}
A level with this deception is created using two types of entities, called obstacle and obstructed. The obstacle blocks the OSS of the obstructed entity. The agent should first substantially collapse the obstacle to clear a path to reach the obstructed entity. To ensure the agent interacts with the obstacle with the objective of collapsing to clear the path, entities that can only be collapsed with a specific strategy are selected as obstacles 
Algorithm \ref{alg:algorithm_clearing_paths} shows the pseudocode of this level template.

\begin{algorithm}[t]
\footnotesize
\algsetup{linenosize=\footnotesize}
\caption{Clearing paths}
\label{alg:algorithm_clearing_paths}
\textbf{Input}: Obstacle Entity, Obstructed Entity\\
\textbf{Output}: A level with clearing paths deception

\begin{algorithmic}[1] 
\IF {obstacle entity can be cleared substantially}
\IF {obstructed entity has an OSS}
\WHILE{generation unsuccessful}
\STATE generate level by matching obstacle and obstructed entities
\STATE verify reachability of targets
\IF {generation successful}
\STATE \textbf{break}
\ENDIF
\IF {maximum generation attempts reached}
\STATE \textbf{return} none
\ENDIF
\ENDWHILE
\STATE allocate birds to level
\STATE generate solution strategy
\ENDIF
\ENDIF
\STATE \textbf{return} generated level
\end{algorithmic}
\end{algorithm}

\subsubsection{Entity Strength Analysis}

\begin{algorithm}[t]
\footnotesize
\algsetup{linenosize=\footnotesize}
\caption{Entity strength analysis}
\label{alg:algorithm_entity_strength}
\textbf{Input}: Entity 1, Entity 2\\
\textbf{Output}: A level with entity strength analysis deception

\begin{algorithmic}[1] 
\FOR{birdX in all bird types}
\FOR{birdY in all bird types except BirdX}

\IF {entity1 solvable by birdX and entity2 solvable by birdY}
\IF {entity1 birdX usage $<$ entity2 birdX usage $\algorithmicand$ entity2 birdY usage $<$ entity1 birdY usage}
\STATE generate level with entity1, entity2, birdX, birdY
\STATE \textbf{break}
\ENDIF

\ELSIF{entity1 solvable by birdY and entity2 solvable by birdX}
\IF {entity1 birdY usage $<$ entity2 birdY usage $\algorithmicand$ entity2 birdX usage $<$ entity1 birdX usage}
\STATE generate level with entity1, entity2, birdX, birdY
\STATE \textbf{break}
\ENDIF
\ENDIF
\ENDFOR
\ENDFOR

\IF {generation successful}
\STATE verify reachability of targets
\STATE generate solution strategy
\STATE \textbf{return} generated level
\ENDIF
\STATE \textbf{return} none
\end{algorithmic}
\end{algorithm}

The strength of an entity is considered with respect to its capability of protecting the pigs within it. In this deceptive level template different bird types are used, as this affects the maximum damage that can be done to an entity. An agent needs to analyze the strength of the entities in the level and determine the correct bird type to use on each entity. 
The number of birds that needs to solve an entity is termed as the \textit{bird usage} in the pseudocode. The template creates levels by selecting two entities: one entity that is strong against one bird type (i.e., bird usage is high) but weak against another (i.e., bird usage is low) and another entity that shows opposite strengths and weaknesses to the same bird types. 
Algorithm \ref{alg:algorithm_entity_strength} shows the pseudocode of this level template.

\subsubsection{Non-greedy Actions}
In Angry Birds, greedy agents tend to kill the most pigs in a single action. In this level template, an entity that has more pigs and easy to solve (termed as a greedy entity) are combined with an entity that has a few pigs and hard to solve (termed as a non-greedy entity). The easiness/hardness of an entity is based on the type and the number of birds needed to solve it. 
To ensure the level can only be solved by doing the non-greedy action first, the two selected entities should have OSSs with different bird types. When generating the level, the order of birds is selected such that the birds needed to solve the non-greedy entity are given first. As the agent cannot change the order of the birds, if it chooses to solve the greedy entity first then it cannot solve the non-greedy entity. 
Algorithm \ref{alg:algorithm_non_greedy_actions} shows the pseudocode of this level template.

\begin{algorithm}[t]
\footnotesize
\algsetup{linenosize=\footnotesize}
\caption{Non-greedy actions}
\label{alg:algorithm_non_greedy_actions}
\textbf{Input}: Greedy Entity, Non-greedy Entity\\
\textbf{Output}: A level with non-greedy actions deception

\begin{algorithmic}[1] 
\IF {2 entities' 2 OSSs use different birds}
\IF {greedy entity is easier to solve than non-greedy entity}
\WHILE{generation unsuccessful}
\STATE generate level with greedy entity and non-greedy entity
\STATE allocate birds in the order for non-greedy and greedy entities
\STATE verify reachability of targets
\IF {generation successful}
\STATE \textbf{break}
\ENDIF
\IF {maximum generation attempts reached}
\STATE \textbf{return} none
\ENDIF
\ENDWHILE
\STATE generate solution strategy
\ENDIF
\ENDIF
\STATE \textbf{return} generated level
\end{algorithmic}
\end{algorithm}

\subsubsection{Non-fixed Tap Times}
The effect of a bird's special power on an entity depends on the time that an agent taps the bird during its flight. A bird can make more damage to the entity if it is tapped at the correct time. The correct tap time needs to be determined considering the distance to the target and the bird's special power. Entities that can only be solved by using a specific tap time of a bird are selected as feasible entities to generate levels with this deception. Two entities with two different tap times are used to generate a level, to ensure that agents with fixed tap times fail to solve the level. 
Algorithm \ref{alg:algorithm_tap_time} shows the pseudocode of this level template.

\begin{algorithm}[t]
\footnotesize
\algsetup{linenosize=\footnotesize}
\caption{Non-fixed tap time}
\label{alg:algorithm_tap_time}
\textbf{Input}: Entity 1, Entity 2\\
\textbf{Output}: A level with non-fixed tap time deception

\begin{algorithmic}[1] 
\FOR{birdX in all bird types with special powers}
\IF {entity1 only solvable by birdX's specific tap time (T\textsubscript{x})}
\FOR{birdY in all bird types with special powers}
\IF {entity2 only solvable by birdY's specific tap time (T\textsubscript{y})}
\IF {T\textsubscript{x} $\neq$ T\textsubscript{y}}
\STATE generate level with entity1, entity2, birdX, birdY
\STATE \textbf{break}
\ENDIF
\ENDIF

\ENDFOR
\ENDIF

\ENDFOR

\IF {generation successful}
\STATE verify reachability of targets
\STATE generate solution strategy
\STATE \textbf{return} generated level
\ENDIF
\STATE \textbf{return} none
\end{algorithmic}
\end{algorithm}

\subsubsection{TNT}
TNT deception is generated using two level templates. Using the first template, if there is an OSS for an entity and that strategy involves targeting TNTs, then a level is generated directly with this entity and the birds need for its OSS. This form of the deception needs an agent to understand that TNTs explode when hit to cause greater damage, which can be used to achieve the goal. In the second template, TNTs are used to distract an agent. This template selects an entity with TNTs and without pigs (termed as a distracting entity) along with another entity with pigs (termed as a distracted entity). The birds allocated to the level are the birds needed for the OSS of the distracted entity. Shooting a bird to explode the TNTs will not help solve the level, and will not leave enough birds to solve the distracted entity. The pseudocode for the second template is shown in Algorithm \ref{alg:TNT}. 

\begin{algorithm}[t]
\footnotesize
\algsetup{linenosize=\footnotesize}
\caption{TNT}
\label{alg:TNT}
\textbf{Input}: Distracting Entity, Distracted Entity\\
\textbf{Output}: A level with TNT deception

\begin{algorithmic}[1] 
\IF {distracting entity has TNTs and no pigs}
\IF {distracted entity has an OSS}
\STATE generate level with distracting entity and distracted entity
\STATE allocate birds needed for the distracted entity's OSS
\STATE verify reachability of targets
\STATE generate solution strategy
\ENDIF
\ENDIF
\STATE \textbf{return} generated level
\end{algorithmic}
\end{algorithm}

Once a deceptive level and its solution have been generated using one of these templates, they are passed to the \textit{Level Validation Module}.

\subsection{Level Validation Module}
The last module in the generation process validates the physical stability and solvability of the generated levels. 
In Angry Birds, all the objects should remain stationary at the start of the level. The \textit{Level Validation Module} first verifies the stability of the level using the Box2D physics engine. The velocities of objects in the level are observed after two seconds of simulating the level to determine the stability of the level.
After confirming stability, the \textit{Level Validation Module} then verifies the solvability of the level, using its generated solution strategy. The final outcome of this process is a stable deceptive level, along with its confirmed solution.

\section{Results and Evaluations} \label{experiments}

\begin{figure*}

  \begin{subfigure}[b]{0.49\columnwidth}
    \includegraphics[width=\linewidth]{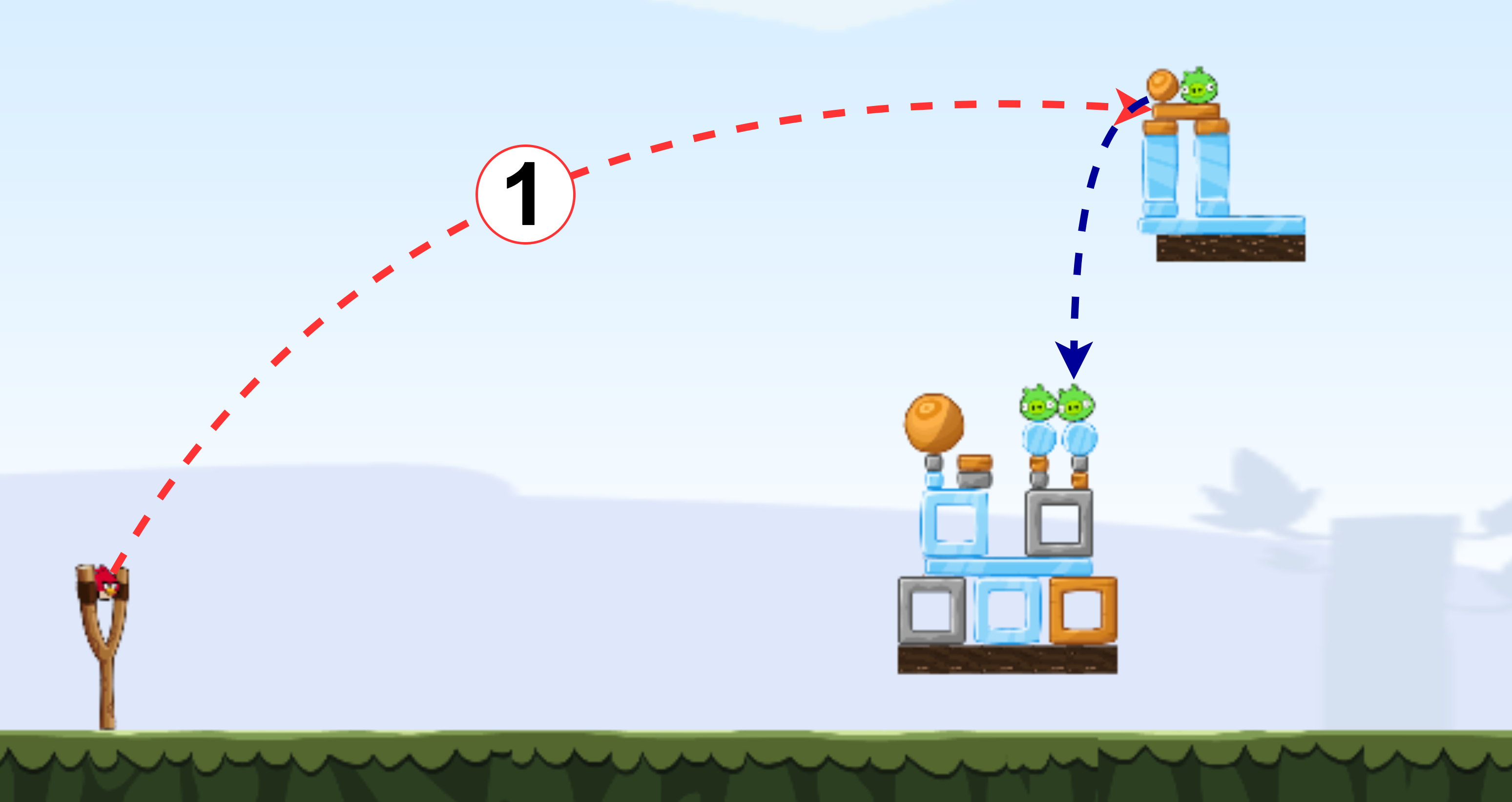}
    \caption{Rolling/Falling Objects}
    \label{rolling_with_sol}
  \end{subfigure}
  \hfill 
  \begin{subfigure}[b]{0.49\columnwidth}
    \includegraphics[width=\linewidth]{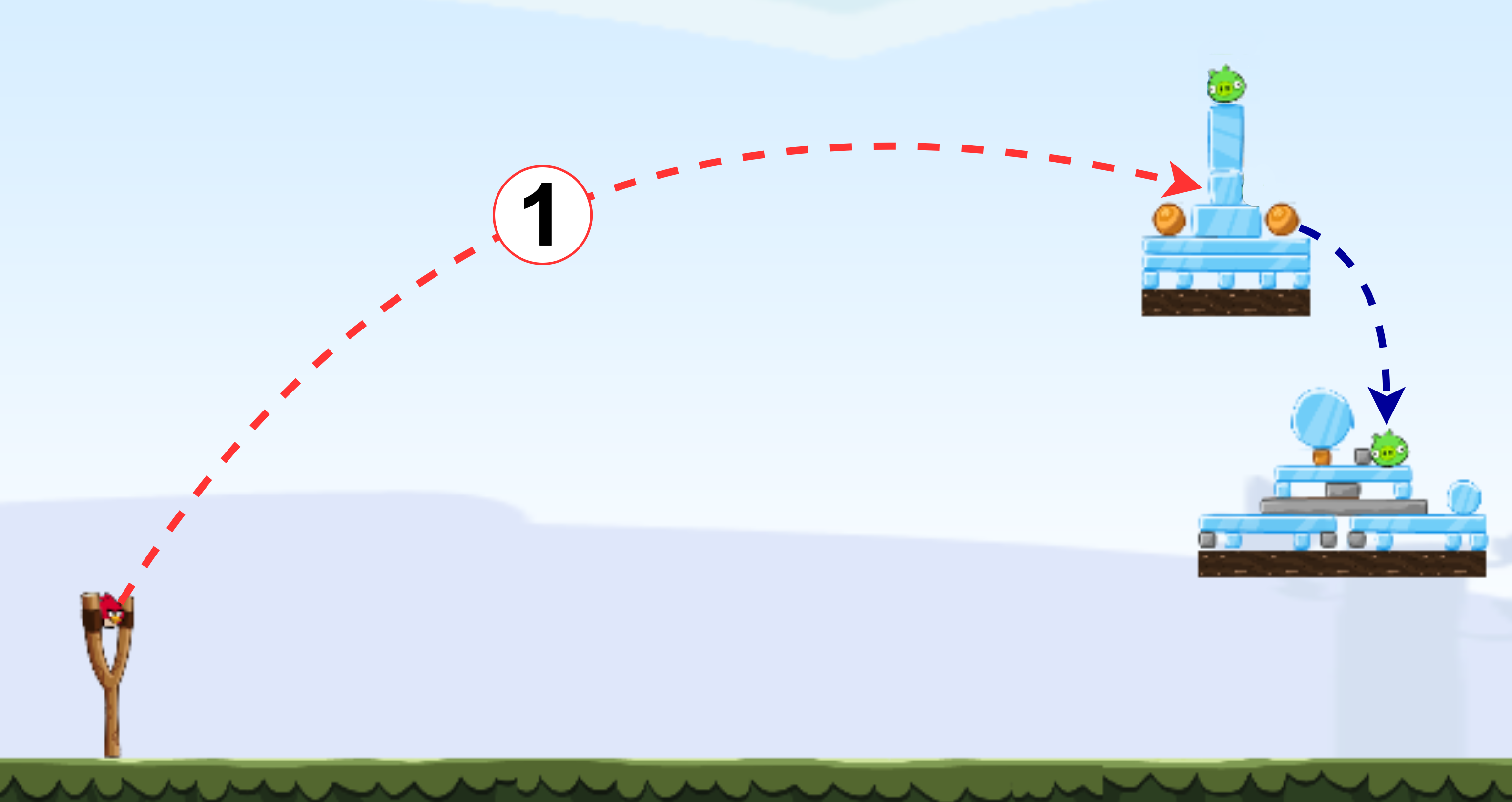}
    \caption{Rolling/Falling Objects}
    \label{clearing_with_sol}
  \end{subfigure}
    \begin{subfigure}[b]{0.49\columnwidth}
    \includegraphics[width=\linewidth]{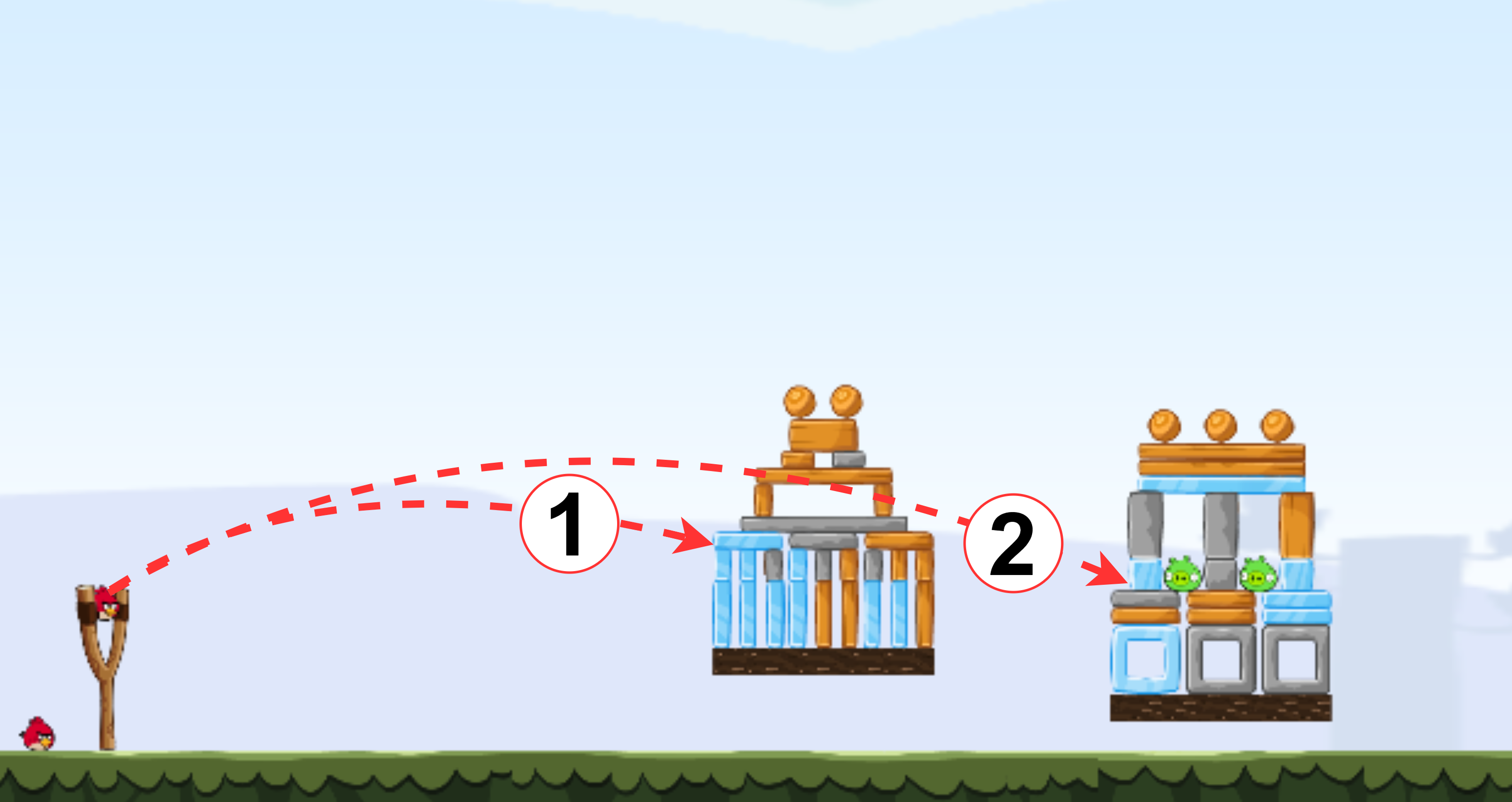}
    \caption{Clearing Paths}
    \label{strength_with_sol}
  \end{subfigure}
  \hfill 
  \begin{subfigure}[b]{0.49\columnwidth}
    \includegraphics[width=\linewidth]{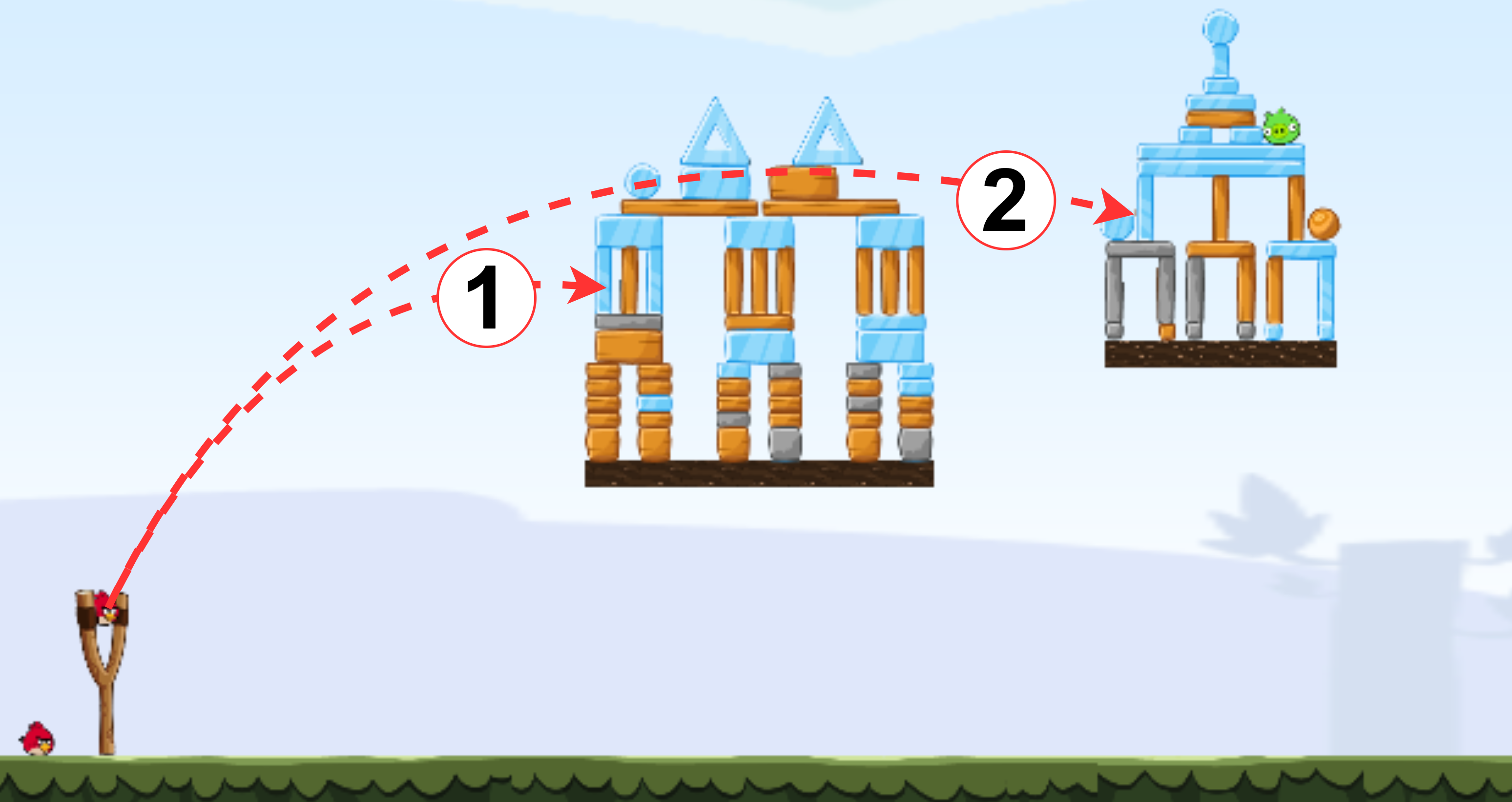}
    \caption{Clearing Paths}
    \label{greedy_with_sol}
  \end{subfigure}
    \begin{subfigure}[b]{0.49\columnwidth}
    \includegraphics[width=\linewidth]{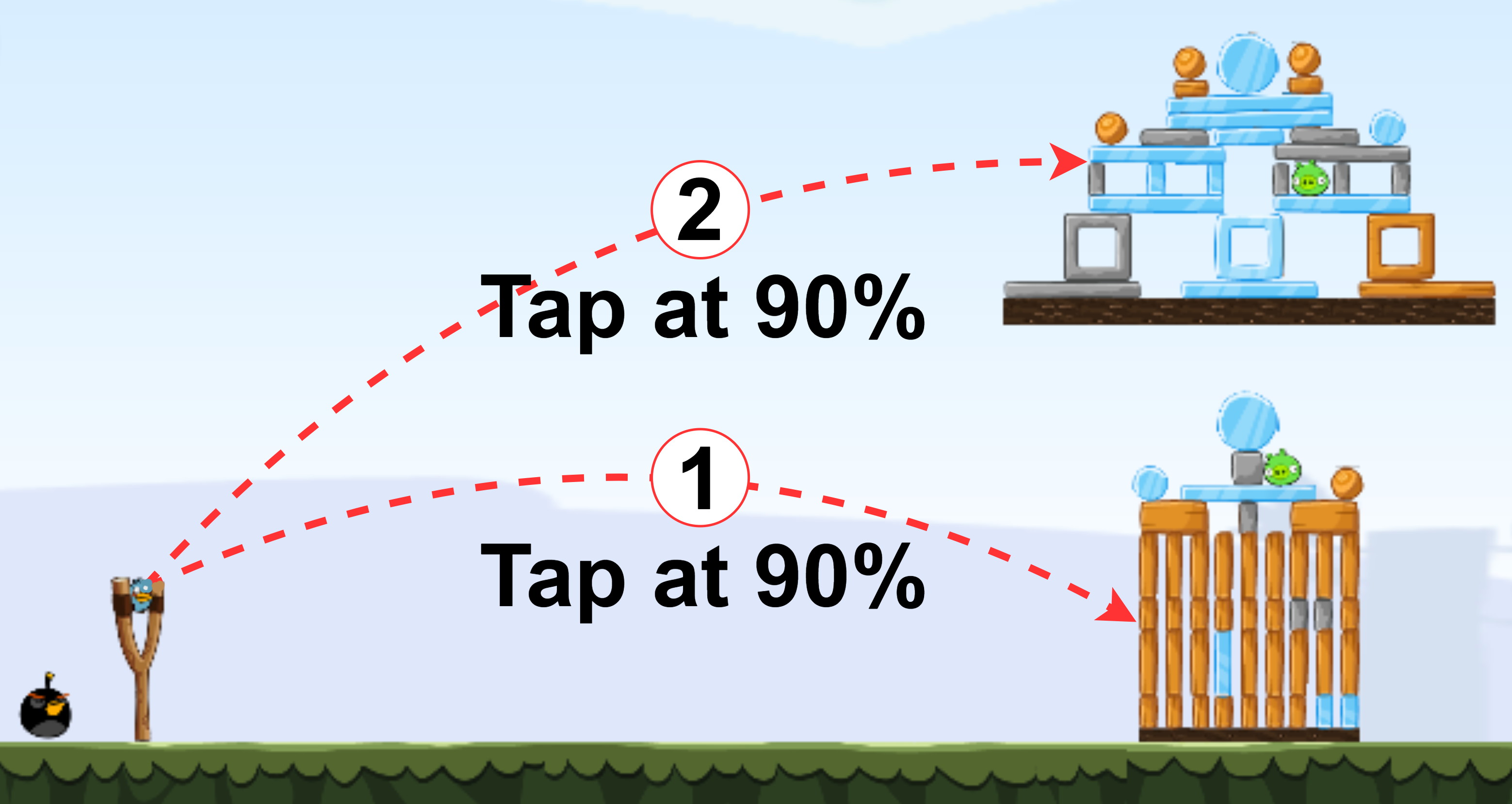}
    \caption{Entity Strength Analysis}
    \label{rolling_with_sol}
  \end{subfigure}
  \hfill 
  \begin{subfigure}[b]{0.49\columnwidth}
    \includegraphics[width=\linewidth]{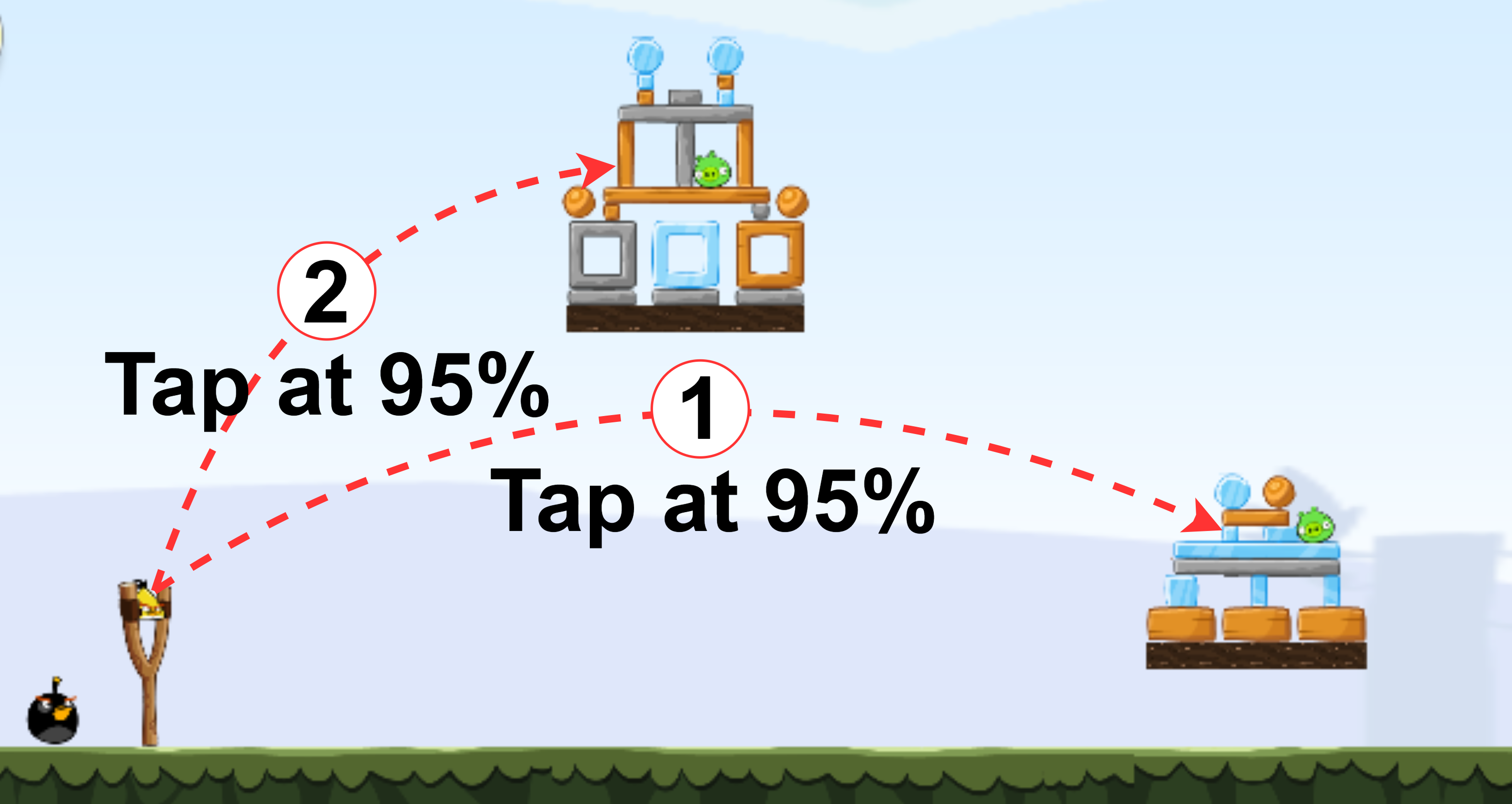}
    \caption{Entity Strength Analysis}
    \label{clearing_with_sol}
  \end{subfigure}
    \begin{subfigure}[b]{0.49\columnwidth}
    \includegraphics[width=\linewidth]{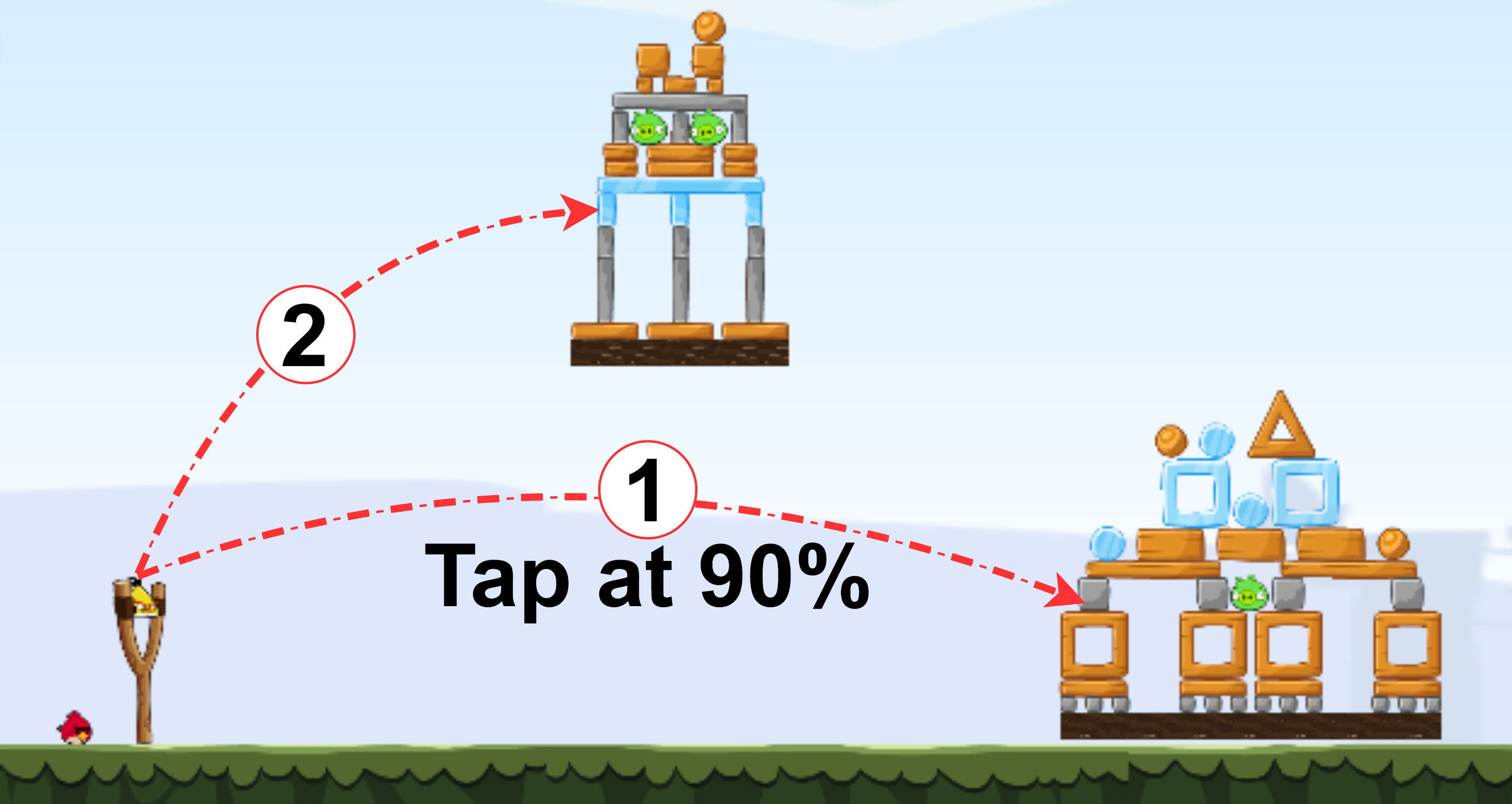}
    \caption{Non-Greedy Actions}
    \label{strength_with_sol}
  \end{subfigure}
  \hfill 
  \begin{subfigure}[b]{0.49\columnwidth}
    \includegraphics[width=\linewidth]{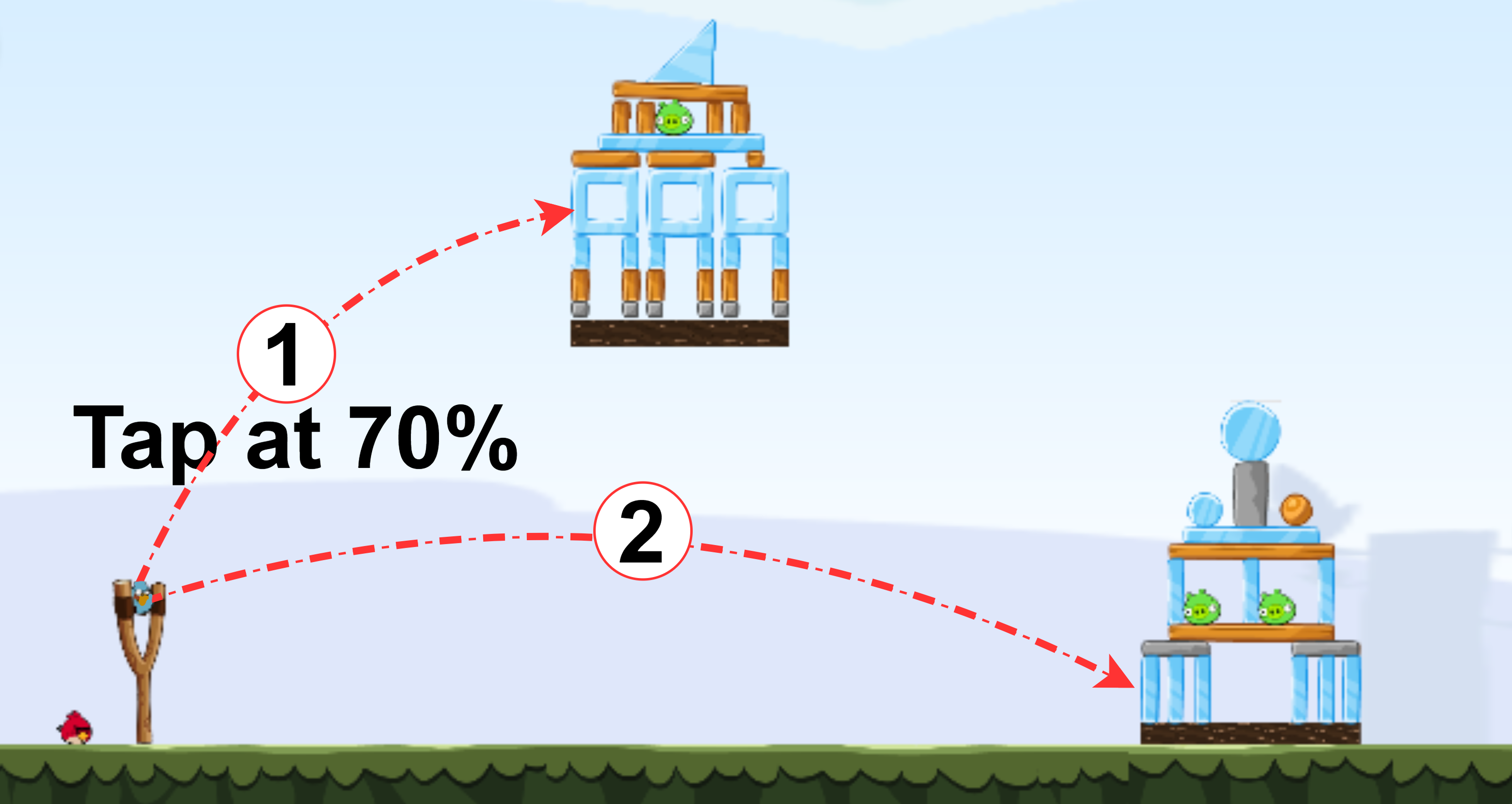}
    \caption{Non-Greedy Actions}
    \label{greedy_with_sol}
  \end{subfigure}
  
    \begin{subfigure}[b]{0.49\columnwidth}
    \includegraphics[width=\linewidth]{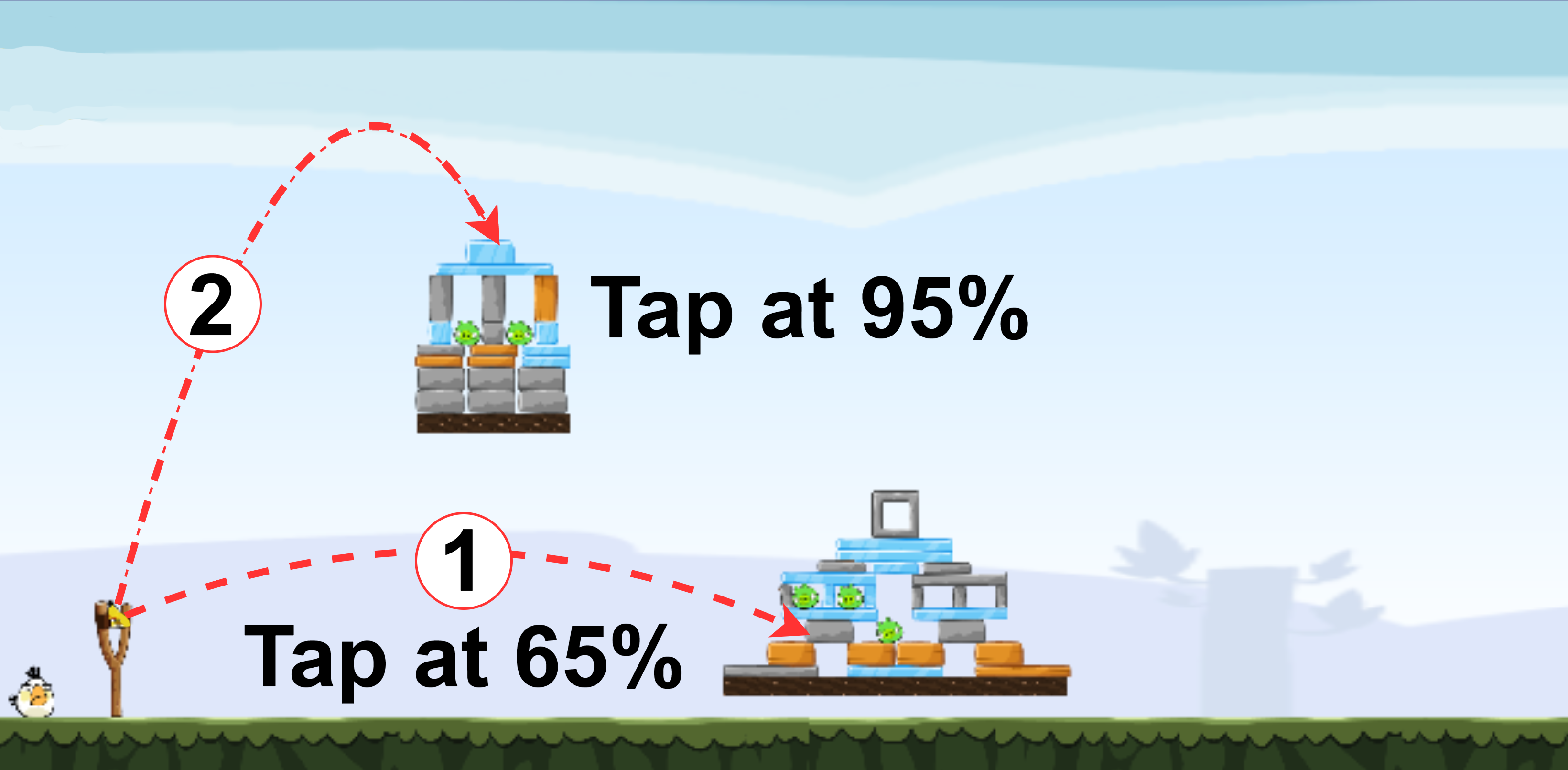}
    \caption{Non-fixed Tap Time}
    \label{rolling_with_sol}
  \end{subfigure}
  \hfill 
  \begin{subfigure}[b]{0.49\columnwidth}
    \includegraphics[width=\linewidth]{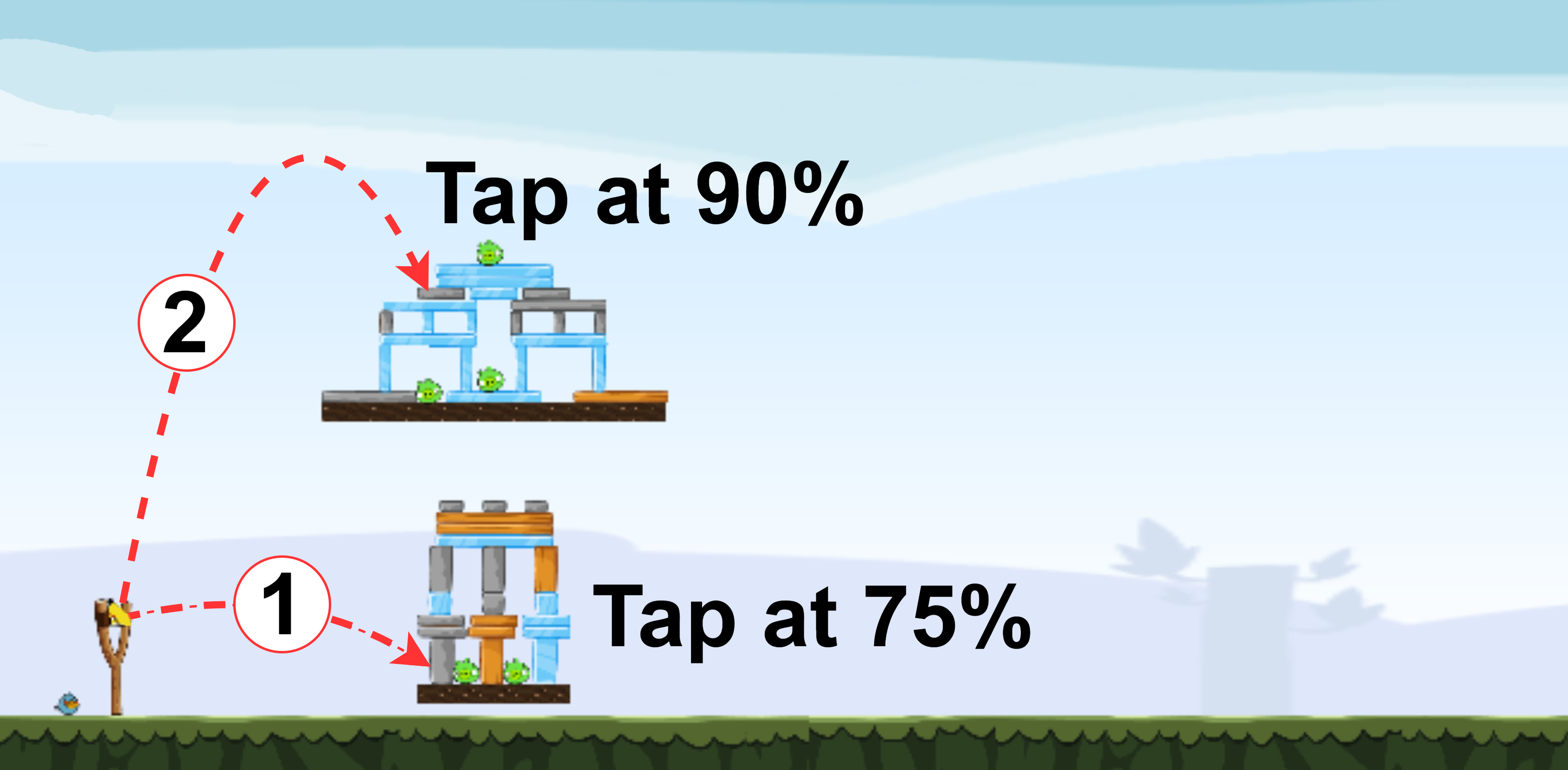}
    \caption{Non-fixed Tap Time}
    \label{clearing_with_sol}
  \end{subfigure}
    \begin{subfigure}[b]{0.49\columnwidth}
    \includegraphics[width=\linewidth]{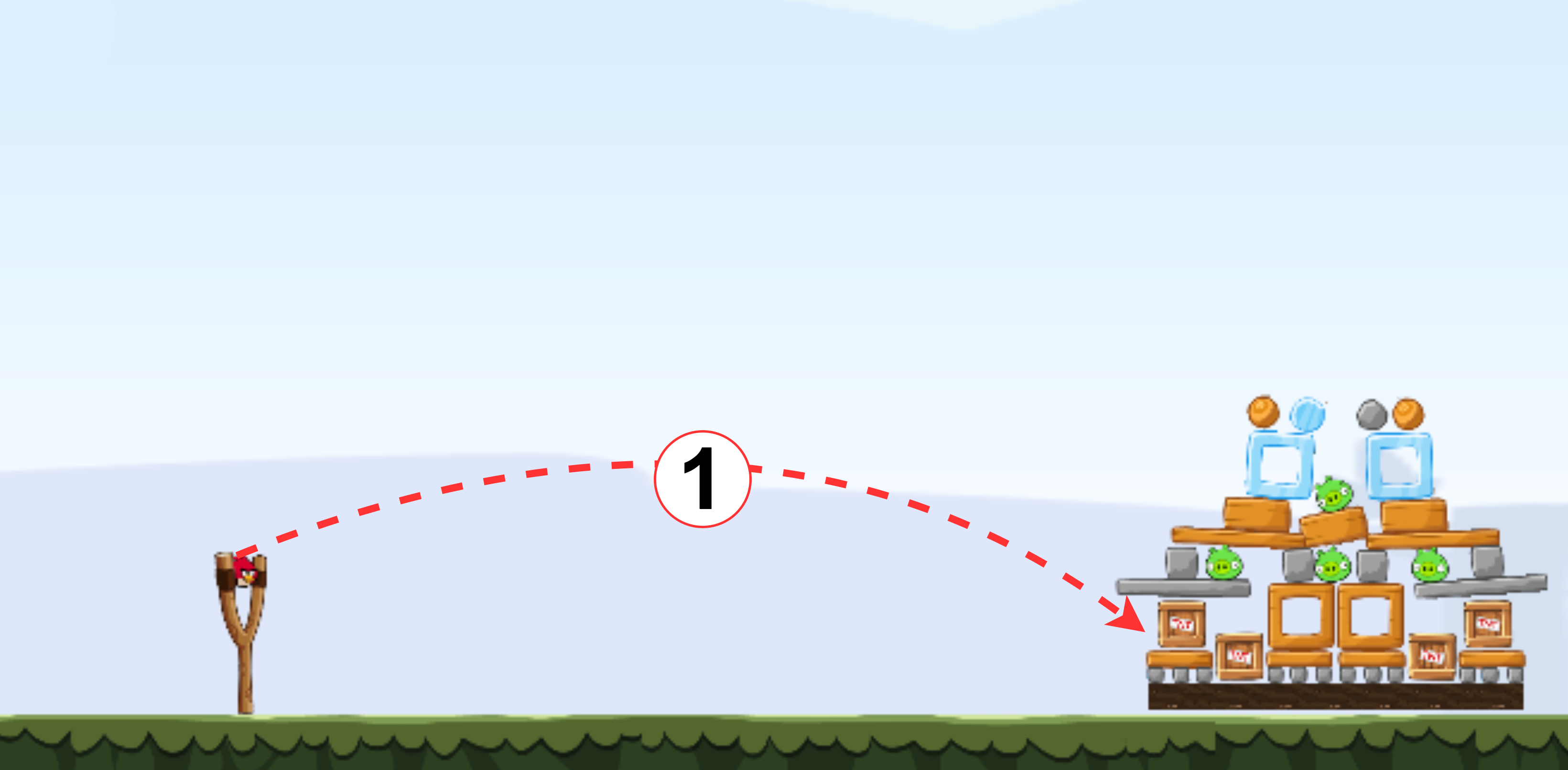}
    \caption{TNT (Solution)}
    \label{strength_with_sol}
  \end{subfigure}
  \hfill 
  \begin{subfigure}[b]{0.49\columnwidth}
    \includegraphics[width=\linewidth]{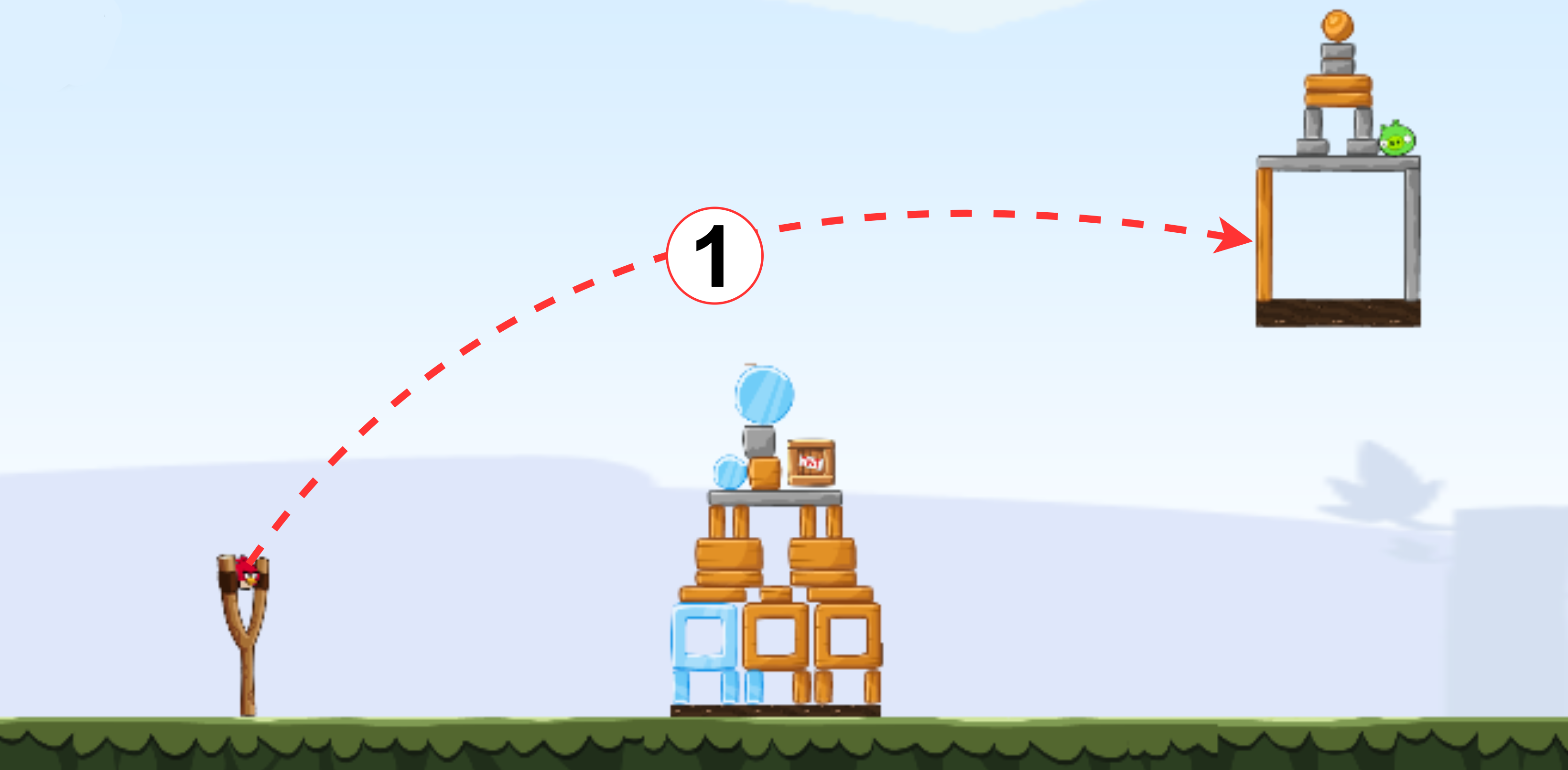}
    \caption{TNT (Distraction)}
    \label{greedy_with_sol}
  \end{subfigure}
  
\caption{Examples of generated levels for the six deception categories. The solution strategy for each level is illustrated using the red dotted arrows. The arrow path shows the trajectory of the bird and the arrowhead is pointed to the target. Numbers show the order of bird shots. ``Tap at $x\%$" means the bird is needed to tap at the length of $x\%$ in its trajectory path, which is only available for birds with special powers. The blue lines in (a) and (b) are the trajectories of falling objects.}
\label{examples_for_4_deceptions}
\end{figure*}

In this section, we present the results and evaluation of the proposed deceptive level generation procedure. The majority of our implementation was coded using Python 3.7. The exceptions were the \textit{Physical Entity Analyzer Module} and the \textit{Level Validation Module} that required simulations done using the Box2D physics engine in Unity, coded in C\#. The simulations were done by speeding up the physics engine by 50 times. The software was run on Windows 10 desktop PC with an i9-9900KS CPU and 64GB RAM. The average time consumed by each module is: \textit{Physical Entity Extraction Module} took 21 milliseconds to extract an entity with 16 blocks, \textit{Physical Entity Analyzer Module} took 85.71 seconds to analyze an entity using 10 strategies, \textit{Level Template Matching Module} took 2.64 seconds to generate a level, and \textit{Level Validation Module} took 9.01 seconds to validate a level. Therefore, the online stage of the generation process can generate a deceptive level in 11.65 seconds on average. Twelve levels generated for the six deception categories are shown in Fig. \ref{examples_for_4_deceptions}.

In the following sections we evaluate our generated levels using four metrics that measure the stability, solvability, deceptiveness, and similarity to human-created levels. 
Even though both stability and solvability are normally verified by the \textit{Level Validation Module}, this module was disabled for these experiments in order to assess how often a generated level passed each of these validity checks.

\begin{table}[t]
\caption{Stability rate ($R_{i}$), solvability rate ($S_{i}$), the average deceptive score ($D_{i}$), and the average difference of the solve rates for human-created and generated levels ($C_{i}$) for the six deception categories.}
\label{table:solvability_deceptiveness}
\begin{center}
\begin{tabular}{|c|c|c|c|c|}
\hline
\textbf{Deception Category} & \textbf{$R_{i}$} & \textbf{$S_{i}$} & \textbf{$D_{i}$} & \textbf{$C_{i}(\%)$} \\
\hline
Rolling/falling Objects & 1.00 & 0.82 & 0.91 & 3.41\\
Clearing Paths & 0.98 & 0.95 & 0.90 & -1.75\\
Entity Strength Analysis & 0.98 & 0.90 & 0.89 & -6.67\\
Non-greedy Actions & 1.00 & 0.81 & 0.95 & 7.56\\
Non-fixed Tap Time & 0.97 & 0.87 & 0.97 & 0.00\\
TNT & 0.96 & 0.92 & 0.95 & -9.26\\
\hline
\end{tabular}
\label{tab1}
\end{center}
\end{table}

\subsection{Stability}
The stability rate ($R_i$) of a deception category $i$ is calculated as the percentage of levels which were stationary within the Box2D physics engine. A set of 100 levels for each deception category was used (600 levels in total). The results are presented in the second column of Table \ref{table:solvability_deceptiveness}. This shows that our generation procedure is capable of generating levels with a stability rate of almost 100\%.

\subsection{Solvability}

The physical outcomes of the interactions considered in the generation process might differ from the outcomes when actually playing the level, typically due to differences in the location of entities when generating levels versus their original location during entity analysis, as discussed in Section \ref{entity_analyzer_module}. Therefore, a generated level might not always be solvable using the intended solution strategy (i.e., the generated solution). Hence, we evaluate the solvability of the generated levels to measure the competence of the generator in creating levels that can be solved with the intended solution strategy. 

The solvability rate ($S_i$) of a deception category $i$ is calculated as the percentage of levels which could be solved using its provided solution strategy. A set of 50 levels for each deception category was used (300 levels in total). The results are presented in the third column of Table \ref{table:solvability_deceptiveness}. This shows that our generation procedure is capable of generating levels with a solvability rate higher than 81\% for the six deceptions.




\subsection{Deceptiveness} 
To measure the degree of deception of the levels in a deception category we define a metric, average deceptive score. A level is considered less deceptive if it can be solved by multiple strategies without understanding the deception. The solution strategy created when generating a level is the strategy that an agent would follow if the deception is understood. To calculate the deceptive score of a level, the level is tested with different known strategies and the number of strategies that solved the level, excluding the solution strategy, is used. Equation \ref{avg_deceptive_score} shows the average deceptive score ($D_i$) calculated for a deception category $i$ by averaging the deceptive scores of all the levels tested for the deception. $N$ is the total number of levels tested, $T_n$ is the total number of strategies used to test the $n^{th}$ level, and $P_{nt}$ is 1 if the $n^{th}$ level is solved by $t^{th}$ strategy or 0 otherwise. $D_i\in [0,1]$ and a higher score means a higher deceptiveness.

\begin{equation}
\label{avg_deceptive_score}
\resizebox{0.5\linewidth}{!}{$
D_i = \dfrac{1}{N}\sum_{n=1}^N \dfrac{1}{T_n} \sum_{t=1}^{T_{n}} (1 - P_{nt})
$}
\end{equation}

For this experiment, we developed a portfolio agent with 10 variants of tactics mentioned in Section \ref{entity_analyzer_module}. A set of 50 levels for each deception category was used (300 levels in total). The fourth column of Table \ref{table:solvability_deceptiveness} shows the average deceptive score calculated for the six deceptions. The results show that the proposed method can generate deceptive levels with an average deceptive score over 0.89 for the six deceptions. 



\subsection{Comparison with Human-created Levels}

From human capabilities, we are very adept at creating deceptive levels. Therefore, we evaluate the generated levels against human-created levels by examining whether the generated levels exhibit similar characteristics to the human-created levels for AI agents. We compare the solve rates of the agents for human-created and generated levels that belong to the same deception category. The average level solve rate difference between the human-created and generated levels ($C_i$) for a deception category $i$ is shown in Equation \ref{avg_passing_rate_difference}. $A$ is the number of agents tested, $M$ is the total number of human-created levels, and $N$ is the total number of generated levels. $P_{am}$ is 1 if the $a^{th}$ agent solved $m^{th}$ human-created level or 0 otherwise, similarly $P_{an}$ is 1 if the $a^{th}$ agent solved $n^{th}$ generated level or 0 otherwise. A positive value for $C_i$ indicates that the generated levels are more difficult to solve than the human-created levels on average for AI agents and vice versa.

\begin{equation}
\label{avg_passing_rate_difference}
\resizebox{0.7\linewidth}{!}{$
C_i = \dfrac{1}{A}\sum_{a=1}^A \left( \dfrac{1}{M} \sum_{m=1}^{M} P_{am} - \dfrac{1}{N} \sum_{n=1}^{N} P_{an} \right)
$}
\end{equation}

\begin{figure}[t]
\centerline{\includegraphics[width=0.5\textwidth]{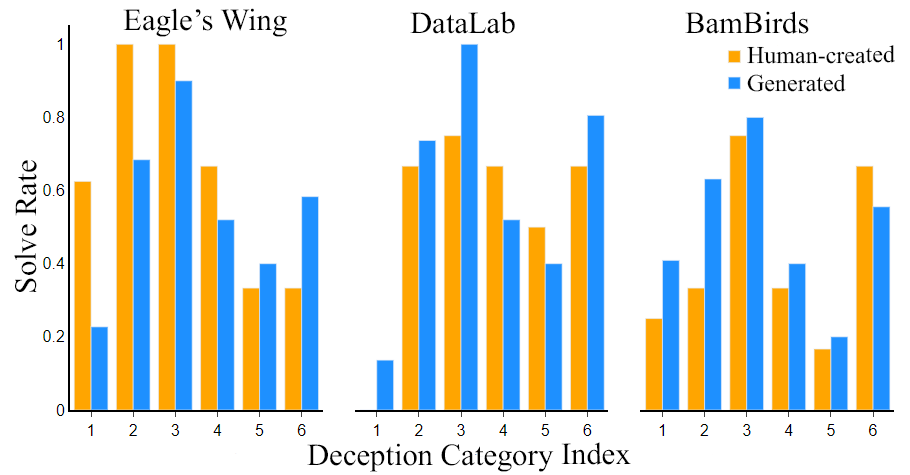}}
\caption{Level solve rates of three agents for human-created and generated levels. The deception indexes on the x-axis are in the order: rolling/falling objects (1), clearing paths (2), entity strength analysis (3), non-greedy actions (4), non-fixed tap time (5), and TNT (6).}
\label{generated_handcraft_comparison}
\end{figure}

For this evaluation, we used 30 handcrafted deceptive levels for Angry Birds from the previous work \cite{deceptive_angry_birds} representing the six deception categories (five levels on average per deception category). We generated 300 levels from our generator (50 levels per deception category). Three state-of-the-art Angry Birds agents Datalab, Eagle’s Wing, and Bambirds from previous AIBirds competitions \cite{stephenson20182017} were used to run the experiment. The solve rates of the three agents for the six deceptions for human-created and generated levels are shown in Fig. \ref{generated_handcraft_comparison}. This figure depicts that the solve rates of the generated levels are correlated to the solve rates of the human-created levels. This portrays that agents show similar behaviours when playing both human-created and generated levels. The $C_i$ values calculated for the six deceptions using the results of the three agents are in the fifth column of Table \ref{table:solvability_deceptiveness}. The levels generated for two deception categories were more difficult to solve than the human-created levels while three deception categories were easier. The human-created and generated levels of TNT deception were equally difficult for the agents to solve. 

\section{Conclusions and Future Work}
In this paper, we have presented a methodology to generate deceptive game levels for Angry Birds. The proposed methodology can generate levels for six deception categories that the state-of-the-art Angry Birds playing agents are vulnerable to. Even though the idea of handcrafting deceptive levels for Angry Birds has been previously investigated, this is the first attempt at generating deceptive levels for a complex physics-based game like Angry Birds. In addition, our approach generates the solution for the levels, which is a feature that is not available in any of the existing Angry Birds level generators, and which could be beneficial for learning agents in their training process. The levels generated from the proposed procedure was evaluated using four metrics: stability, solvability, deceptiveness, and a comparison with human-created levels. The results of these metrics demonstrate that the generation process can competently create deceptive levels for the six deceptions considered.

This work aims to enable the development of advanced Angry Birds playing agents that can perform well under deceptions by providing sufficient training/testing data. To deal with these deceptions, the AI techniques of the current agents can be improved to expand their reasoning, planning, and generalizations skills. Future level generation research can involve generating more complex levels by combining multiple deception categories. Additional deception categories may also be proposed, if more flaws within the state-of-the-art agents can be identified. The deception categories we considered, rolling/falling objects, clearing paths, entity strength analysis, and non-greedy actions can also be seen in a real physical environment. Therefore, this research can serve as a base for generating deceptive tasks in a real physical environment.

\bibliographystyle{ieeetr}

\bibliography{references}

\end{document}